\documentclass[sigconf]{acmart}
\AtBeginDocument{%
  }

\acmYear{2025}

\acmConference[Workshop on AI and Data Science for Digital Finance held in conjunction with  ICAIF 2025]{International Conference on AI in Finance}{November  15-18,
  2025}{Singapore}

\usepackage{graphicx}
\usepackage{booktabs}
\usepackage{amsmath}
\usepackage{multirow}
\usepackage{bm}

\title{How Data Quality Affects Machine Learning Models for Credit Risk Assessment}

\author{Andrea Maurino}
\affiliation{
  \institution{University of Milano-Bicocca}
  \country{Italy}
}
\email{andrea.maurino@unimib.it}

\begin{document}

\begin{abstract}
Machine Learning (ML) models are being increasingly employed for credit risk evaluation, with their effectiveness largely hinging on the quality of the input data. In this paper we  investigate the impact of several data quality issues, including missing values, noisy attributes, outliers, and label errors,  on the predictive accuracy of the machine learning model used in credit risk assessment. Utilizing an open-source dataset, we introduce controlled data corruption using the Pucktrick library to assess the robustness of 10 frequently used models like Random Forest, SVM, and Logistic Regression and so on. Our experiments show significant differences in model robustness based on the nature and severity of the data degradation. Moreover, the proposed methodology and accompanying tools offer practical support for practitioners seeking to enhance data pipeline robustness, and provide researchers with a flexible framework for further experimentation in data-centric AI contexts.
\end{abstract}




\maketitle

\section{Introduction}

Financial stability represents a foundational requirement for modern banking systems, underpinning sustained economic development. Among the various indicators of financial soundness, credit risk plays a central role. It reflects the possibility that a borrower will fail to meet their financial obligations, and it can stem from defaults, excessive exposure to specific sectors or individuals, or sovereign inability or unwillingness to repay debt. As such, credit risk evaluation constitutes a central task in banking, encompassing all stages of the loan lifecycle, from origination to repayment and recovery.

Over the past two decades, the increasing complexity of global markets, combined with events such as the 2007–2008 financial crisis and the introduction of Basel regulatory frameworks, has led to a growing emphasis on the need for accurate and transparent credit risk models. To mitigate the likelihood of loan defaults and minimize the accumulation of Non-Performing Assets (NPAs), financial institutions have progressively adopted machine learning (ML) techniques to improve credit scoring and risk evaluation. While ML models have demonstrated strong predictive capabilities, their effectiveness heavily depends on the quality of the data used during training—a factor that is often assumed rather than examined in depth.

In this paper, we examine the impact of training data quality on the effectiveness of credit risk models. This study is prompted by the observation that while contemporary ML methods have shown efficacy in creditworthiness prediction, their reliance on the quality of data has not been thoroughly examined\cite{zhang2024mlfinance}.

We use an open source credit risk dataset\footnote{\url{https://www.kaggle.com/datasets/laotse/credit-risk-dataset}} as a stand-in for real credit data and apply a set of controlled data manipulations via the Pucktrick library to simulate realistic degradations.   
In addition to the core analysis, this paper proposes an empirical methodology to simulate data imperfections and assess their impact on predictive performance. We also describe the main tools used to support the development of this study.

The objective is to assist practitioners in identifying which ML models are more robust to real-world noise and to highlight essential preprocessing strategies that can mitigate the impact of data quality issues.

Our findings reveal that the introduction of certain types of errors can, counterintuitively, improve model performance—yielding up to a 17\% increase in F1 score for specific models. At the same time, the extensive set of analyses allows us to identify error patterns that significantly degrade performance. An additional insight from our results is that feature-specific errors do not affect all models uniformly; the same corrupted feature may have a negligible impact on one model and a severe impact on another.

Beyond the results obtained on the specific dataset used in this study, the paper offers practical guidelines for enhancing the reliability of predictive models operating in noisy environments.

The remainder of this paper is organized as follows. Section \ref{sec:soa} reviews the state of the art on how data quality affects the performance of financial models. Section \ref{sec:dataset} describes the dataset used and outlines the preprocessing steps applied. Section \ref{sec:meth} details the methodology adopted in the study, while the following section introduces the experimental setup, with a focus on the new features of the PuckTrick library. This library supports the injection of various types of errors according to user-defined error models. Section \ref{sec:result} presents a comprehensive overview of the experimental results. In Section \ref{sec:discussion}, we discuss the main findings, and finally, Section \ref{sec:conclusion} draws conclusions and outlines directions for future works.

\section{Related  works}
\label{sec:soa}
\subsection{Errors in credit risk modelling}
Several papers explored the impact of different types of dirty data (among others: inconsistent, noise, duplicate, and outlier) on ML models \cite{Qi2024,10.1007/s00521-022-07702-7, SUN20073358, 10.14778/3648160.3648178}. 
Recently, in \cite{Mohammed_2025}, a systematic and comprehensive empirical study was presented to investigate the impact of various data quality dimensions on different machine learning tasks. The study explores several ML algorithms and simulates different data quality scenarios affecting both training and test datasets. Six data quality dimensions are considered: consistent representation, completeness, feature accuracy, target accuracy, uniqueness, and class balance. For each dimension, a dedicated parameterized data polluter is implemented to inject specific types of data errors. No technical details are provided about the design or implementation of these data polluters, 

Several papers propose solutions for specific type of errors in data such as umbalanced data, missing values.  In \cite{CHEN2024357} authors examined the impact of class imbalance on both predictive performance and interpretability in credit scoring.  Starting with a large UK mortgage dataset (2016–2020) and two public credit scoring datasets, the authors progressively increase the imbalance between defaulters and non-defaulters to study its effects. Results shows that severe class imbalance not only degrades the detection of rare default events, but also destabilizes model explanations: popular post-hoc explainers (LIME and SHAP) become significantly less stable as imbalance grows

In \cite{blattner2021costlynoisedatadisparities}, the authors investigate how the precision of credit scores varies across demographic groups, showing that minority and low-income borrowers are assessed with systematically noisier signals of default risk. Using large-scale credit bureau and mortgage datasets, they find that commonly used credit scores (e.g., VantageScore 3.0) exhibit 7–9\% lower predictive performance (AUC) for these underserved groups. Their analysis attributes this performance gap primarily to data bias—such as sparse or “thin” credit files and reduced credit history—rather than to modeling bias stemming from algorithmic limitations.
Authors of \cite{Florez-Lopez01032010} analyzed the impact of missing data on credit risk models, investigating how different types of missing data (MCAR, MAR, and MNAR processes) affect credit risk. Multiple imputation (MI) techniques, which utilize Markov Chain Monte Carlo and resampling methods, were identified as effective solutions for addressing credit risk issues related to missing data.

Authors of \cite{guan2023responsible} show a “human-in-the-loop” knowledge acquisition methodology  to construct fair and compliant rules that could also improve overall performance. The proposed framework, enables a human domain expert to incrementally refine and improve a model trained using machine learning by adding rules based on domain knowledge applied to those cases where the machine learning model fails to fairly and correctly predict the outcome. If all cases processed are monitored in this way then domain knowledge could be used to correct any error made by the ML model. 

Unfortunately, no paper has proposed a sound and practical approach to assess how the different data quality dimensions may impact credit risk models.

\subsection{Data corruption tools}
There are a few examples in the literature of tools that have been developed for corrupting data. 
Bart (Benchmarking Algorithms for data Repairing and Translation)\cite{10.14778/2850578.2850579}  is the first framework for the scalable generation of errors in clean relational databases, offering a high degree of control over the characteristics of the introduced errors. Bart allows users to specify the types and distribution of random errors in the presence of data quality rules. These rules are expressed using a denial constraints language (DCs). BART works on relational database tables only and does not include the evaluation of machine learning tasks. Moreover, it is defined for general-purpose data cleaning task and not for machine learning

Gecko\cite{JUGL2024101846} is a Python library for creating shareable scripts to generate and mutate realistic personal data only. The library is particularly useful for testing record linkage algorithms operating on personal identification data (e.g., given and last name, sex, date and place of birth, residential address) to identify similar records; but it is limited to this task only.

Jenga \cite{schelter2021jenga}  is a Python-based library designed to evaluate the impact of data corruptions on machine learning models. It provides a structured way to introduce various types of errors into datasets and measure their effects on predictive performance. Jenga allows the introduction of different types of data corruptions, missing values, swapped values, scaling errors, and noise. 

Pucktrick\cite{puctrick2025} is another Python library created to support both researchers in the testing of their algorithms and practitioners in the testing of their pipelines. It provides a broader range of error types such as mislabel, duplicate, outlier, noise, missing data. 

\section{Dataset and data preparation} 
\label{sec:dataset}
To study how errors in training data affect credit risk modeling we use the credit risk dataset \footnote{\url{https://www.kaggle.com/datasets/laotse/credit-risk-dataset}}. The dataset is composed by 12 columns described in Table \ref{tab:columns} and contains 32.581 rows. 

\begin{table}[h]
    \centering
    \begin{tabular}{|c|p{4cm}|}
        \hline
        \textbf{Name of the Feature} & \textbf{Description} \\
        \hline
        person\_age & Age of the individual applying for the loan. \\
        \hline
        person\_income & Annual income of the individual. \\
        \hline
        person\_home\_ownership & Type of home ownership of the individual. \\
        \hline
        person\_emp\_length & Employment length of the individual in years. \\
        \hline
        loan\_intent & The intent behind the loan application. \\
        \hline
        loan\_grade & The grade assigned to the loan based on the creditworthiness of the borrower. 
        A means the borrower has a high creditworthiness, indicating low risk;  G means the borrower's creditworthiness is the lowest, signifying the highest risk. \\
        \hline
        loan\_amnt & The loan amount requested by the individual. \\
        \hline
        loan\_int\_rate & The interest rate associated with the loan. \\
        \hline
        loan\_percent\_income & The percentage of income represented by the loan amount. \\
        \hline
        cb\_person\_default\_on\_file & Historical default of the individual as per credit bureau records: Y - The individual has a history of defaults on their credit file. N -  The individual does not have any history of defaults. \\
        \hline
        cb\_preson\_cred\_hist\_length & The length of credit history for the individual. \\
        \hline
        loan\_status & Loan status, where 0 indicates non-default and 1 indicates default. \\
        \hline
    \end{tabular}
    \caption{Feature names and descriptions}
    \label{tab:columns}
\end{table}

\begin{figure}
    \centering
    \includegraphics[width=1\linewidth]{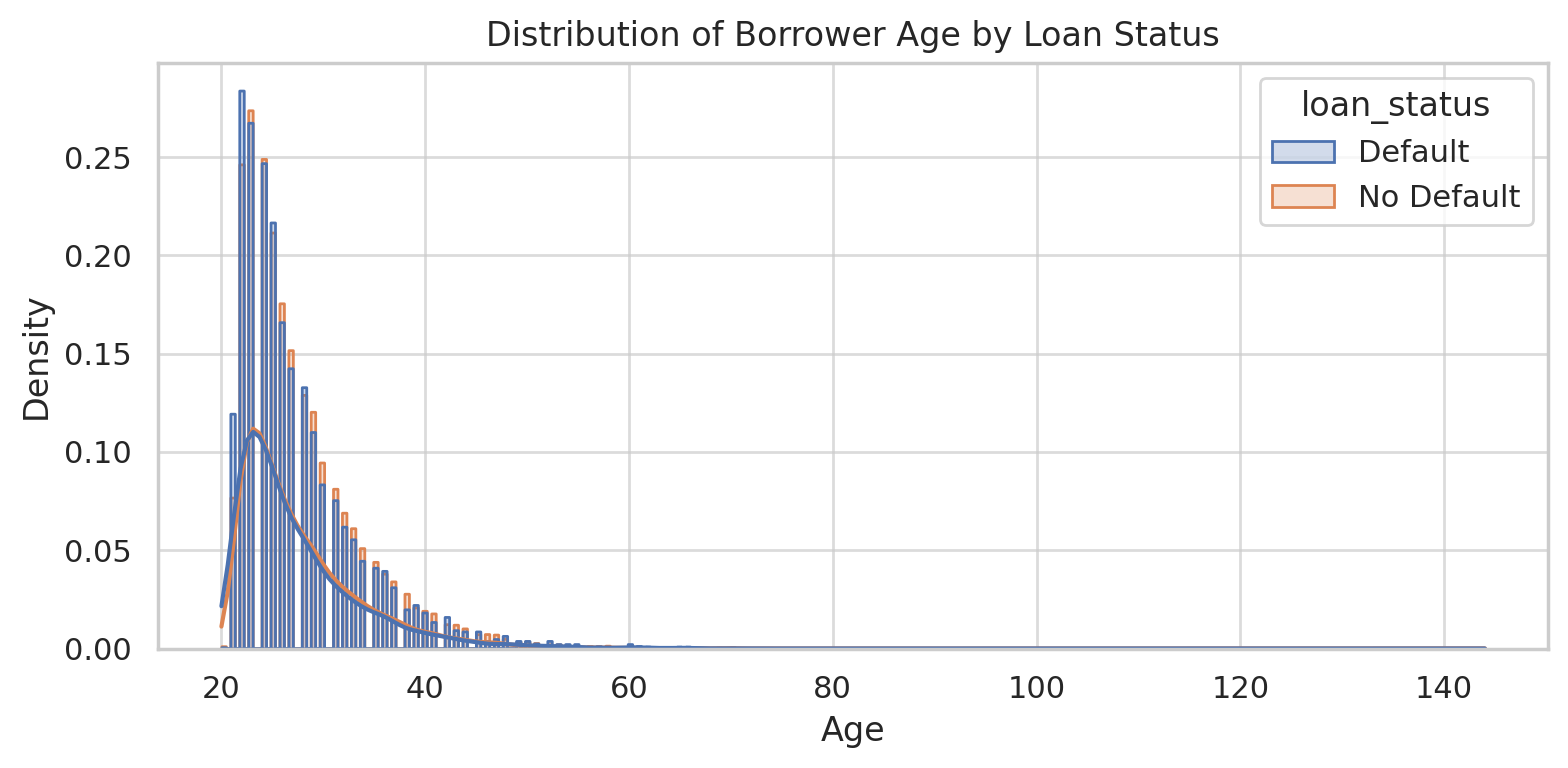}
    \caption{Distribution of Borrower Age by Loan Status}
    \label{fig:age_status}
\end{figure}

The dataset is unbalanced. 25473 rows have \textit{loan\_status} equal to 0 (no-default 78,2\%) while only 7.108 rows are related to default clients (21,8\%). The original dataset includes 165 duplicated rows. 
By analysing the relationship between age and the target variable (see figure \ref{fig:age_status}) it is possible to say that age may have moderate discriminative power. While both defaulting and non-defaulting borrowers span a similar age range, the density curves indicate subtle differences in the distribution, particularly among younger individuals.

The probability of default appears to be slightly higher among borrowers in their early 20s. In contrast, borrowers aged 30 and above show a relatively lower and more stable default rate. This suggests that younger age groups may represent a higher-risk segment.

Given the continuous nature of the variable and the observed concentration of defaults in specific age bands, it may be beneficial to apply transformations such as binning (e.g., into age groups) or normalization. These approaches could help capture non-linear effects and stabilize the influence of age in models sensitive to feature scale.

\begin{figure}
    \centering
    \includegraphics[width=1.2\linewidth]{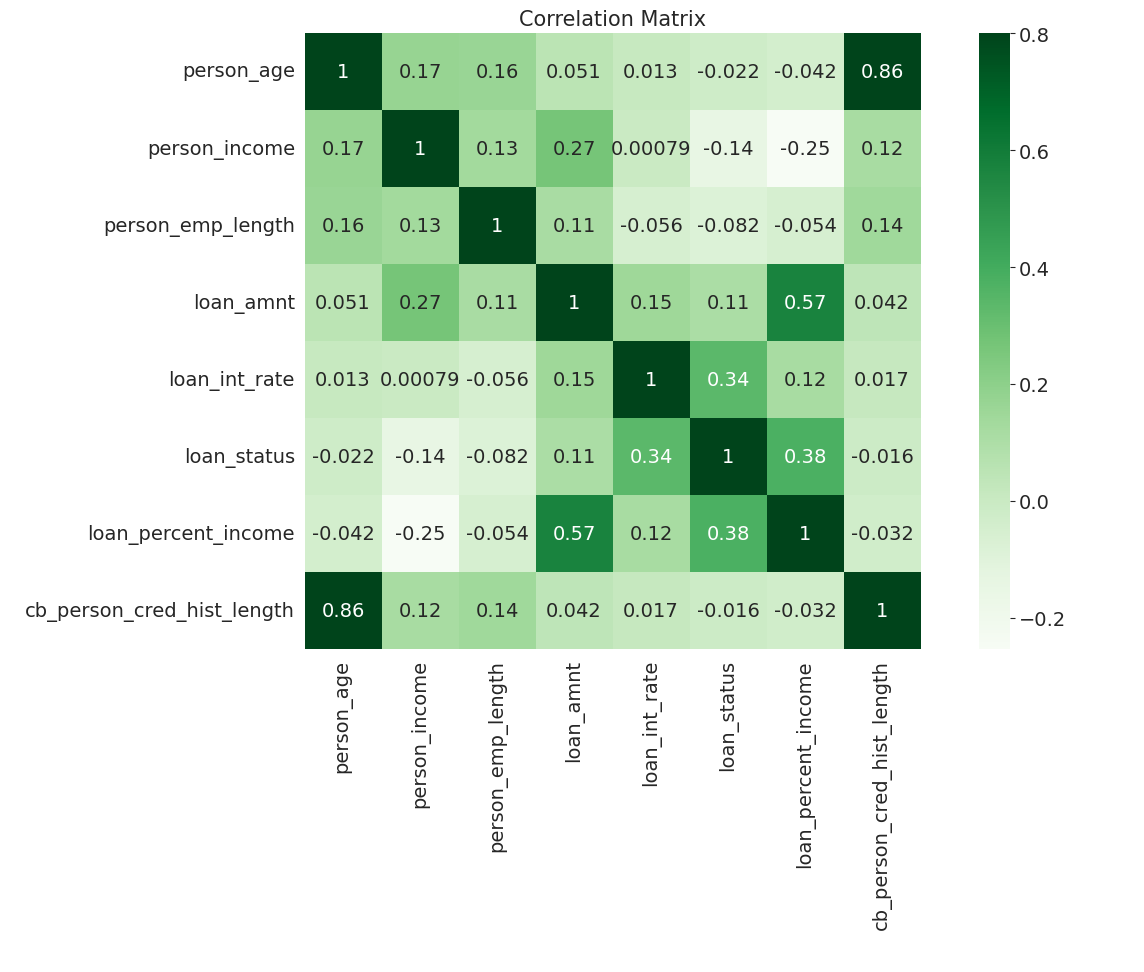}
    \caption{Correlation matrix}
    \label{fig:correlation}
\end{figure}

The correlation matrix (Figure \ref{fig:correlation}) shows that the target variable \textit{loan\_status}  has only weak correlations with the other features. The strongest correlation is with \textit{loan\_int\_rate} (0.34) and \textit{loan\_percent\_income} (0.38), suggesting that higher interest rates and a larger portion of income allocated to the loan are slightly associated with higher default risk. Two variables \textit{cb\_person\_cred\_hist\_length} and \textit{person\_age} show a significant correlation

Before applying machine learning models, we conducted standard preprocessing operations, such as eliminating rows with null values (comprising less than 10\% of the dataset) and creating bins for variables including \textit{person\_age} as previously described, \textit{income}, and \textit{loan\_amount}.  The final training dataset includes 14 features, 1 target variable,  and 28.637 rows 

\section{Methodology}
\label{sec:meth}
\begin{figure}[h]
    \centering
    \includegraphics[width=0.5\textwidth]{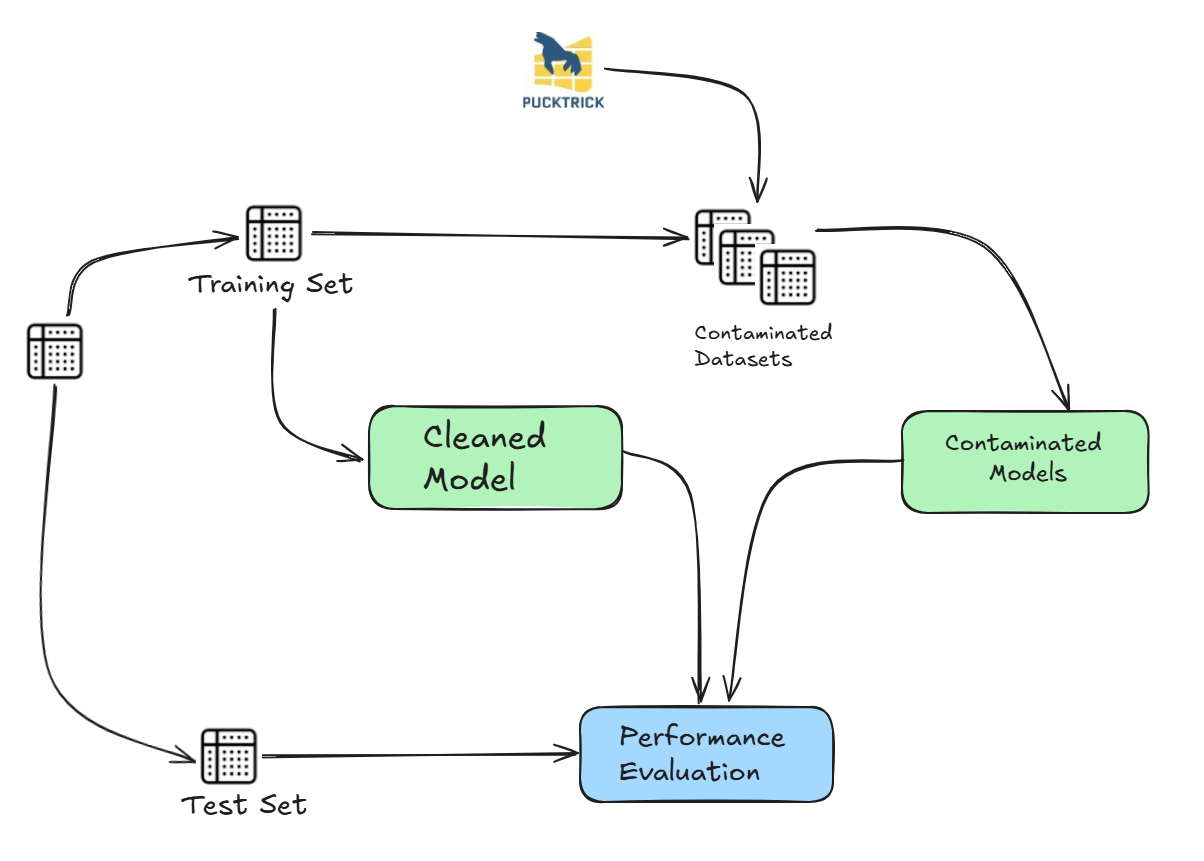}
    \caption{Methodological Workflow}
    \label{fig:methodology}
\end{figure}

To rigorously assess the impact of various data imperfections on the performance of different machine learning algorithms applied to credit risk, a specific methodology was defined as shown in Figure \ref{fig:methodology}.
The selected data undergoes a structured process of preparation and analysis. Specifically, the dataset is first split into a training set and a test set, ensuring a proper division for model evaluation. The training set is then deliberately "dirtied" using the Pucktrick library, allowing for controlled corruption of the data.\\

Once the data corruption process is complete, different machine learning models are applied to both the training and test sets. However, for the training set, models are tested on both the original clean data and the artificially corrupted data, enabling a comparative analysis of their behavior under different conditions.  \\

Multiple evaluation metrics were employed in the performance evaluation phase. These metrics offer complementary insights into classification quality, allowing for a comprehensive understanding of how each model responds to different sources of noise and degradation in the input data. By analyzing these metrics, we can quantify the robustness and generalization ability of the classifiers under non-ideal conditions.\\
Accuracy measures the overall proportion of correctly classified instances out of all instances:

\[
\text{Accuracy} = \frac{TP + TN}{TP + TN + FP + FN}
\]

where \(TP\) (True Positives) and \(TN\) (True Negatives) are the counts of correctly predicted positive and negative instances, respectively, while \(FP\) (False Positives) and \(FN\) (False Negatives) denote the counts of incorrect predictions.

Precision quantifies the proportion of correctly predicted positive instances among all instances predicted as positive:

\[
\text{Precision} = \frac{TP}{TP + FP}
\]

Recall (also called Sensitivity or True Positive Rate) measures the proportion of actual positive instances correctly identified:

\[
\text{Recall} = \frac{TP}{TP + FN}
\]

The F1 score is the harmonic mean of precision and recall, balancing their trade-offs:

\[
F1 = 2 \times \frac{\text{Precision} \times \text{Recall}}{\text{Precision} + \text{Recall}}
\]

\section{Experimental Setup}
\label{sec:experiment}

According to the above mentioned methodology, we corrupt the credit risk dataset by means of Pucktrick.
The library implements error injections across five key categories \cite{puctrick2025}

\begin{itemize}
\item \textbf{Duplicates}: insertion of copies of existing rows within the dataset, including not only exact duplicates but also modified duplicates created by altering values in selected columns.

\item \textbf{Noise}: introduction of noise into data,  by replacing values with numbers, date, category uniformly sampled within features original range. In case of strings feature, random strings are generated. 

\item \textbf{Outliers}: addition of anomalous values applied not only to continuous features but also to various data types, allowing the simulation of atypical or extreme values across heterogeneous columns. Values are with the range $[3*|\sigma|,5*|\sigma|]$ for higher values or  $[-5*|\sigma|,-3*|  \sigma|]$ for lower values. 
\item \textbf{Missing Values}: controlled removal of data points by setting values to \textit{NaN}.
\item \textbf{Wrong Labels}: intentional corruption or alteration of target labels by swapping values among class .
\end{itemize}

The library is designed to be used in both experimental research scenarios focused on data quality and robustness, and practical applications such as stress testing ML pipelines in realistic, error-prone environments.

In the new version of the library, used in this paper, we introduced the concept of  \textbf{error model}, that is, a formal specification of how imperfections or corruptions are introduced into a clean dataset. It defines the nature, scope, and behavior of the errors being injected, and serves as a conceptual framework for simulating realistic data degradation scenarios.

The error model implemented in Pucktrick is designed to be \textit{modular}, \textit{configurable}, and \textit{reproducible}. It treats each type of data imperfection — such as duplicates, outliers, noise, missing values, or wrong labels — as a parameterized transformation applied to a defined subset of data, according to user-defined error models.
An error model is defined as 6-npla $\mathbb{E}= ( \mathbb{D},\varepsilon,\sigma,\rho,p,\varphi)$:\\

Where:
\begin{itemize}
    \item $\mathbb{D}=(F_1,F_2,\dots,F_n)$ is the tabular dataset to be contaminated composed by features $F_1,F_2,\dots,F_n$
    \item $\varepsilon$ is the type of error to insert
    \item $\sigma$ is the set of features $F_i\subseteq \mathbb{D}$ to be modified together
    \item $\rho$ is the set of predicate for defining a subset of the $\mathbb{D} $ to be modified
    \item p is the percentage of $\rho$ to be modified
    \item $\eta$ is the distribution of probability function used for inserting errors. Examples of functions are normal, gaussian, Poisson function. 
    \item $\varphi$ is the corruption mode; and may assume one of the following two values: \texttt{new} (fresh injection) or \texttt{extended} (incremental contamination).
\end{itemize}

By abstracting each corruption process as a configurable, repeatable transformation, Pucktrick enables the construction of realistic and controlled error scenarios that can be consistently applied across different datasets and experiments. This level of control is crucial for studying the behavior of machine learning models under degraded data conditions and comparing the robustness of alternative algorithms or pipelines.

In the credit risk analysis, we defined the following error models:
\begin{enumerate}
    
    \item $\mathbb{E}_1=Labels,\sigma=Loan\_status,\rho=\mathbb{D},p=0.3|0.5, \eta=normal, \phi=new|extended $
    \item $\mathbb{E}_1=Duplicate,\sigma=(F_1,F_2,\dots,F_n) \subseteq \mathbb{D},\rho=\mathbb{D},p=0.3|0.5, \eta=normal, \phi=new|extended $
     \item $\mathbb{E}_1=Missing,\sigma=F_1|F_2|\dots|F_n ,\rho=\mathbb{D},p=0.3|0.5, \eta=normal, \phi=new|extended $
     \item $\mathbb{E}_1=Noise,\sigma=F_1|F_2|\dots|F_n ,\rho=\mathbb{D},p=0.3|0.5, \eta=normal, \phi=new|extended $
     \item $\mathbb{E}_1=Outlier,\sigma=F_1|F_2|\dots|F_n ,\rho=\mathbb{D},p=0.3|0.5, \eta=normal, \phi=new|extended $
\end{enumerate}
Error Model $\mathbb{E}_1$ applies the label error to the target variable \textit{Loan\_status} to the entire dataset. The model apply two levels of corruption: 30\% with a $\phi$ equal to new and 50\% with a $\phi$ equal to extended. Error Model $\mathbb{E}_2$ introduces corruption across all features of the dataset simultaneously. Error Models $\mathbb{E}_3$,$\mathbb{E}_4$, and $\mathbb{E}_5$ each inject errors into a single feature at a time, resulting in 14 distinct datasets per error model. In total, this process yields more than 80 corrupted datasets\footnote{cleaned dataset, all corrupted ones and the ML models results are available \url{https://anonymous.4open.science/r/icaif25-4235/}}.
\\

In the experiments we conducted we used the 10 well-knonwn classifiers:
\begin{itemize}
    \item {Linear Discriminant Analysis (lda)}
    \item {Logistic Regression (lr)}
    \item {Extra Trees Classifier (et)}
    \item {Random Forest (rf)}
    \item {K-Nearest Neighbors (knn)}
    \item {Multi-Layer Perceptron (mlp)}
    \item {Naive Bayes (nb)}
    \item {Decision Tree (dt)}
    \item {Support Vector Machine (svm)}
    \item {Quadratic Discriminant Analysis (qda)}
\end{itemize}

Models are trained on contaminated training sets and evaluated on clean test data using an 80/20 split. Performance is measured using the F1 Score.
We selected contamination rates of 30\% and 50\% to represent moderate and severe degradation scenarios, respectively, while keeping the analysis concise and within space constraints. After injecting controlled errors into the datasets using predefined perturbation strategies, a series of machine learning models were trained and evaluated through the \texttt{PyCaret}\footnote{\url{https://pycaret.org/}} framework, an open source low-code machine learning library in Python that automates machine learning workflows. In each experiment, models were trained on the corrupted version of the dataset, reflecting the presence of specific types and levels of errors, and evaluated on a clean synthetic test set. This setup was intentionally chosen to assess the robustness of each model: training on noisy data simulates real-world imperfections, while testing on clean data isolates the effect of training-time corruption without introducing further variability at test time.

This approach allows for a precise evaluation of a model’s ability to generalize from imperfect data to ideal conditions, which is crucial in scenarios where models are deployed in controlled environments but trained on historical or potentially flawed datasets. To ensure comparability, all models were trained using PyCaret’s default hyperparameters, avoiding the additional variability that hyperparameter optimization could introduce.

\section{Experimental Results}
\label{sec:result}
Table \ref{tab:model_performance} shows the performance of 10 adopted ML models wrt. the cleaned dataset. Due to the unbalanced nature of the dataset, we used the F1 Score as the main adopted metric. In the following subsections, we report the evolution of performance of selected ML models when different error strategies were applied    

\begin{table}[h]
    \centering
    \begin{tabular}{cccccc}
        \hline
         Model & Accuracy & Precision & Recall & F1\_Score \\
        \hline
         lda & \textbf{0.9354} & \textbf{0.9690} & 0.7192 & \textbf{0.8256} \\
         et & 0.8821 & 0.7129 & \textbf{0.7463} & 0.7292 \\
         lr & 0.9219 & 0.9011 & 0.7110 & 0.7949 \\
         rf & 0.8680 & 0.7296 & 0.6026 & 0.6601 \\
         knn & 0.8533 & 0.7345 & 0.4860 & 0.5850 \\
         mlp & 0.8551 & 0.6535 & 0.6782 & 0.6656 \\
         nb & 0.8247 & 0.7131 & 0.2939 & 0.4163 \\
         dt & 0.8200 & 0.7221 & 0.2496 & 0.3710 \\
         svm & 0.7997 & 0.7290 & 0.0928 & 0.1646 \\
        \hline
    \end{tabular}
    \caption{Performance Metrics of ML Models trained on cleaned dataset}
    \label{tab:model_performance}
\end{table}
\subsection{Misslabel}
\label{sec:label}
Labeling datasets is a critical yet error-prone stage of the data preprocessing pipeline, often affected by ambiguity, annotation bias, or inconsistencies \cite{geiger2021garbage}. Several factors may result in a loan request being incorrectly marked as an NPA even when it has been fully repaid. The F1 Score calculated over the dataset with a 30\%  or 50\% of label swapped is reported in Table \ref{tab:f1_scores_labels}. Downward arrows signify that the F1 score is reduced compared to the model trained on the cleaned dataset. Conversely, upward arrows indicate that training on dirty data resulted in an improved F1 Score.
\begin{table}[h]
    \centering
    \begin{tabular}{lcccc}
        \hline
        \textbf{Model} & \textbf{F1\_Score@30\%} & \textbf{Gain}& \textbf{F1\_Score@50\%}& \textbf{Gain}\\
        \hline
        lda   & 0.6237 &$\downarrow$&0.2360&$\downarrow$\\
        et    & 0.5950 &$\downarrow$&0.000&$\downarrow$\\
        lr    & 0.5577 &$\downarrow$&0.0013&$\downarrow$\\
        rf    & 0.5512 &$\downarrow$&0.2120&$\downarrow$\\
        knn   & 0.4687 &$\downarrow$&0.2938&$\downarrow$\\
        mlp   & 0.4739 &$\downarrow$&0.2959&$\downarrow$\\
        nb    & 0.0039 &$\downarrow$&0.2999&$\downarrow$\\
        dt    & 0.3663 &$\downarrow$&0.2101&$\downarrow$\\
        svm   & 0.0003 &$\downarrow$&0.00036&$\downarrow$\\
        \hline
    \end{tabular}
    \caption{F1 Scores of ML Models in Presence of Mislabeling Errors}
    \label{tab:f1_scores_labels}
\end{table}

It is worth noting that no ML model, trained on contaminated data,  performs better than the corresponding model that was trained on the cleaned dataset. This means that this type of error has a huge impact on the performance of the trained model as already demonstrate in the literature \cite{NEURIPS2022_e7a217c3}

\subsection{Duplicated}
Duplicate row errors occur when identical records appear more than once within a dataset. These errors often stem from flawed data integration processes, repeated data ingestion, or missing primary key constraints. In the financial sector, such anomalies can arise when consolidating client data from multiple systems, especially in mergers or cross-departmental reporting. For example, if a client’s transaction history is collected from both a trading and a CRM platform without proper deduplication, the same transaction might be logged multiple times. 
The experimental results are somewhat surprising and are presented in Table \ref{tab:f1_scores}. An analysis of the data reveals that introducing 30\% duplicated rows led to a notable improvement in the performance of all models considered  with the exception of the Random Forest model. This phenomenon may be attributed to the concept of generalization, that is the model’s ability to learn patterns from the training data and apply them effectively to previously unseen instances, even when these deviate from the original distribution.
\begin{table}[h]
    \centering
    \begin{tabular}{lcccc}
        \hline
        \textbf{Model} & \textbf{F1\_Score@30\%} & \textbf{Gain} &\textbf{F1\_Score@50\%} &\textbf{Gain} \\
        \hline
        lda & 0.9626 &$\uparrow$& 0.9675 &$\uparrow$\\
        et  & 0.9396 &$\uparrow $&0.9443&$\uparrow$\\
        lr  & 0.9554 &$\uparrow$& 0.9621&$\uparrow$\\
        rf  & 0.6593 &$\downarrow $&0.6600&$\downarrow$\\
        knn & 0.6023 &$\uparrow $&0.5874&$\uparrow$\\
        mlp & 0.6774 &$\uparrow $&0.6633&$\downarrow$\\
        nb  & 0.4437 &$\uparrow $&0.4509&$\uparrow$\\
        dt  & 0.5514 &$ \uparrow$& 0.4921&$\uparrow$\\
        svm & 0.5000 &$\uparrow$ & 0.6196&$\uparrow$\\
        \hline
    \end{tabular}
    \caption{F1 Scores of ML Models when duplicate data are at 30\% and 50\%}
    \label{tab:f1_scores}
\end{table}
When the level of duplication is increased to 50\%, we observe that nearly all models exhibit improved performance, with the exception of the Random Forest and MLP models, which show slight performance degradations.

\subsection{Missing Data}
Missing data refers to the absence of values for certain variables in a dataset, which can compromise the validity of statistical analyses and machine learning models. In the financial domain, missing values frequently arise due to system integration issues, delayed reporting, or client non-disclosure. For instance, credit applications may lack complete income or employment history, especially for self-employed individuals. Similarly, transactional data might be incomplete due to outages in data pipelines or failures in third-party reporting services.

\begin{table}[h]
    \centering
    \begin{tabular}{lcccc}
        \hline
        \textbf{Model} &$\overline{\bm{\mathrm{F1\_Score@50}}}$& \textbf{Gain}&$\overline{\bm{\mathrm{F1\_Score@50}}}$& \textbf{Gain}\\
        \hline
        lda & 0.8815 &$\uparrow$&0.8762&$\uparrow$\\
        et  & 0.8935 &$\uparrow$&0.8700&$\uparrow$\\
        lr  & 0.8415 &$\uparrow$&0.8317&$\uparrow$\\
        rf  & 0.6676 &$\uparrow$&0.6830&$\uparrow$\\
        knn & 0.6386 &$\uparrow$&0.6368&$\uparrow$\\
        mlp & 0.0003 &$\downarrow$&0.0287&$\downarrow$\\
        nb  & 0.4102 &$\downarrow$&0.3838&$\downarrow$\\
        dt  & 0.4015 &$\uparrow$&0.4099&$\uparrow$\\
        svm & 0.3986 &$\uparrow$&0.4566&$\uparrow$\\
        \hline
    \end{tabular}
    \caption{Average F1 Scores for  Models Effected by Missing Data}
    \label{tab:avg_f1_scores}
\end{table}

From the examination of Table \ref{tab:avg_f1_scores}, it becomes clear that, on average, all models, except for the Naive Bayes and MLP classifiers, show improvements in performance, some quite significant, when trained on datasets with 30\% and 50\% corruption levels. This suggests that, unlike other models, Naive Bayes and MLP are less tolerant to this type of data corruption, as their performance either remains unchanged or deteriorates. In Table \ref{tab:model_f1_scores}, further insights are provided, illustrating the average F1 Score yielded by each ML model when the feature \textit{loan\_percent\_income}  is corrupted.  Table suggests that no single feature predominantly influences the decline in performance. In other words, corrupting any specific attribute may contribute similarly to the overall decrease in model performance or increase. 
\begin{table}[h]
    \centering
    \begin{tabular}{lcccc}
        \hline
        \textbf{Model} & \textbf{F1 Score@30} &\textbf{Gain}&\textbf{F1 Score@50}\textbf{Gain}\\
        \hline
        lda & 0.8823 &$\uparrow$& 0.8756& $\uparrow$\\
        et  & 0.8901 &$\uparrow$& 0.8906& $\uparrow$\\
        lr  & 0.8368 &$\uparrow$& 0.8381& $\uparrow$\\
        rf  & 0.6702 &$\uparrow$& 0.6705& $\uparrow$\\
        knn & 0.6471 &$\uparrow$& 0.6324& $\uparrow$\\
        mlp & 0.0003 & $\downarrow$&0.0003& $\downarrow$\\
        nb  & 0.4080 & $\downarrow$&0.4062& $\downarrow$\\
        dt  & 0.1397 & $\downarrow$&0.3841& $\uparrow$\\
        svm & 0.3560 & $\uparrow$&0.4985& $\uparrow$\\
        \hline
    \end{tabular}
    \caption{Comparison of F1 Scores of Models when \textit{loan\_percent\_income} is Affected by Missing Data}
    \label{tab:model_f1_scores}
\end{table}

\subsection{Outlier}
Outliers are data points that deviate significantly from the majority of observations in a dataset. In financial applications, and particularly in credit risk modeling, outliers can distort model estimates, leading to unreliable predictions and suboptimal risk assessments. These anomalies may represent rare but valid cases, such as exceptionally high income or unusually large loan amounts, or they may result from data entry errors, reporting mistakes, or technical glitches in data collection systems. In the context of credit risk, outliers can artificially inflate or deflate risk scores, causing misclassification of borrower creditworthiness. For example, an outlier in the \textit{person\_income} variable could lead the model to underestimate a borrower’s default probability. Identifying and handling outliers is therefore essential to ensure model robustness and prevent biased decision-making.

\begin{table}[h]
    \centering
    \begin{tabular}{lcccc}
        \hline
          \textbf{Model} &$\overline{\bm{\mathrm{F1\_Score@50}}}$& \textbf{Gain}&$\overline{\bm{\mathrm{F1\_Score@50}}}$& \textbf{Gain}\\
      
        \hline
        lda & 0.9273 &$\uparrow$&0.9117&$\uparrow$\\
        et  & 0.8998 &$\uparrow$&0.8525&$\uparrow$\\
        lr  & 0.9225 &$\uparrow$&0.9108&$\uparrow$\\
        rf  & 0.6538 &$\downarrow$&0.6825&$\uparrow$\\
        knn & 0.5951 &$\uparrow$&0.5868&$\uparrow$\\
        mlp & 0.6535 &$\downarrow$&0.6314&$\downarrow$\\
        nb  & 0.3711 &$\downarrow$&0.3565&$\downarrow$\\
        dt  & 0.4158 &$\uparrow$&0.3899&$\uparrow$\\
        svm & 0.3188 &$\uparrow$&0.4026&$\uparrow$\\
        \hline
    \end{tabular}
    \caption{Average F1 Scores for ML Models Effected by Outlier}
    \label{tab:avg_f1_scores_out}
\end{table}

An analysis of the results reported in Table \ref{tab:avg_f1_scores_out} shows that, with the exception of MLP and Naive Bayes, almost all models exhibit improved performance when trained on data contaminated with outliers. Notably, the SVM model demonstrates a striking increase in F1 score—nearly doubling compared to its performance on the original dataset. This suggests that the injected outliers have influenced the selection of support vectors, effectively altering the decision boundaries (hyperplanes) in a way that enhances the model’s ability to separate the two classes. In this case, the outliers may have acted as informative data points, helping the SVM to better capture the underlying structure of the data. This unexpected improvement underscores the complex interaction between certain types of noise and model-specific learning mechanisms.

Naive Bayes and MLP models tend to be particularly sensitive to the presence of outliers, which can explain the observed degradation in F1 Score when such values are introduced. In the case of Naive Bayes, its reliance on strong distributional assumptions—such as Gaussianity for continuous features—makes it vulnerable to extreme values that distort the estimated class-conditional probabilities. This leads to poorly calibrated decision boundaries and misclassification, especially when the data is imbalanced. For MLPs, the issue stems from their non-linear structure and reliance on gradient-based optimization: outliers can disproportionately influence the loss function, leading to unstable training dynamics, overfitting, or poor generalization. Since the F1 Score balances both precision and recall, any increase in false positives or false negatives due to outliers has a direct and measurable impact on model performance. 

\subsection{Noise}
\label{sec:noise}
Noise in data refers to the presence of errors, inconsistencies, or irrelevant information that can obscure the underlying patterns. In the context of finance, and particularly in credit risk assessment, noisy data can significantly impair the performance of predictive models. It may lead to inaccurate estimation of default probabilities, misclassification of creditworthiness, and ultimately, flawed decision-making.

The results reported in Table \ref{tab:avg_f1_scores-1} are not straightforward to interpret. Only Logistic Regression (LR), Support Vector Machines (SVM), and Extra Trees (ET) consistently improve their F1 scores under both 30\% and 50\% noise levels. In contrast, Random Forest (RF), Naive Bayes (NB), and Multi-Layer Perceptron (MLP) exhibit a systematic decrease in performance across both corruption levels. The behavior of K-Nearest Neighbors (KNN), Linear Discriminant Analysis (LDA), and Decision Trees (DT) is more variable, with performance either improving or deteriorating depending on the specific level of corruption of the dataset.

\begin{table}[h]
    \centering
    \begin{tabular}{lcccc}
        \hline
            \textbf{Model} &$\overline{\bm{\mathrm{F1\_Score@50}}}$& \textbf{Gain}&$\overline{\bm{\mathrm{F1\_Score@50}}}$& \textbf{Gain}\\
      \hline
        lda  & 0.9137 &$\uparrow$& 0.8934&$\uparrow$\\
        et   & 0.8642 &$\uparrow$& 0.8254&$\uparrow$\\
        lr   & 0.9065 &$\uparrow$& 0.8822&$\uparrow$\\
        rf   & 0.6614 &$\uparrow$& 0.6467&$\downarrow$\\
        knn  & 0.5774 &$\downarrow$& 0.5963&$\uparrow$\\
        mlp  & 0.6333 &$\downarrow$& 0.6141&$\downarrow$\\
        nb   & 0.4025 &$\downarrow$& 0.3506&$\downarrow$\\
        dt   & 0.4754 &$\uparrow$& 0.3619&$\downarrow$\\
        svm  & 0.4257 &$\uparrow$& 0.4233&$\uparrow$\\
        \hline
    \end{tabular}
    \caption{Average F1 Score of ML Models in Case of Noisy Datasets}
    \label{tab:avg_f1_scores-1}
\end{table}

In a similar vein, regarding noise-related errors, none of the features consistently affect the performance across all models. As an example, the F1 scores for the variable \textit{person\_income} fluctuate significantly among different models. For 40\% of the models examined, the F1 Score diminishes. This variation indicates that sensitivity to noise is predominantly model-specific and is not uniformly triggered by inaccuracies in single features.
\begin{table}[h]
    \centering
    \begin{tabular}{lcccc}
        \hline
        \textbf{Model} & \textbf{F1\_Score@30}&\textbf{Gain}& \textbf{F1\_Score@50}&\textbf{Gain} \\
        \hline
        lda & 0.9275 &$\uparrow$&0.9145&$\uparrow$\\
        et  & 0.8792 &$\uparrow$&0.8583&$\uparrow$\\
        lr  & 0.9281 &$\uparrow$&0.9266&$\uparrow$\\
        rf  & 0.6600 &$\downarrow$&0.6605&$\uparrow$\\
        knn & 0.6270 &$\uparrow$&0.6263&$\uparrow$\\
        mlp & 0.5627 &$\downarrow$&0.5197&$\downarrow$\\
        nb  & 0.1159 &$\downarrow$&0.1161&$\downarrow$\\
        dt  & 0.3560 &$\downarrow$&0.0051&$\downarrow$\\
        svm & 0.5763 &$\uparrow$&0.3560&$\uparrow$\\
        \hline
    \end{tabular}
    \caption{Comparison of ML Models Affected by Noise Based on F1 Score for the feature \textit{person\_income}}
    \label{tab:noise_1}
\end{table}

\section{Discussion} 
\label{sec:discussion}
The findings of analysis suggest that model selection in credit risk assessment should consider not only accuracy on clean data but also robustness under imperfect conditions. 
First of all, the MLP and Naive Bayes classifiers are the only two models that exhibit a significant decline in performance when errors are introduced into the training data. This suggests that they are particularly sensitive to data imperfections and may lack robustness in noisy environments.
In contrast, for most of the other models, the results indicate that the injection of certain types of errors—such as duplicates and missing values—can unexpectedly enhance predictive performance. This counterintuitive behavior may be due to the regularizing effect that mild perturbations can have on some algorithms, helping to improve generalization or reduce overfitting under specific conditions. As shown in Table \ref{tab:top_model}, the lda model achieves an F1 score of 0.9675 when a 50\% of duplicated is introduced, representing an improvement of 17\%  wrt the model trained on the original, uncorrupted dataset.

Concerning the deployment of pipelines for model retraining or gathering inference data, the changes in model behavior due to noise, as discussed in Section \ref{sec:noise}, emphasize the necessity of diligently tracking such errors during pipeline development. Furthermore, in the phase of retraining models, it is crucial to focus on the labeling process, since Section \ref{sec:label} demonstrated that label noise can considerably degrade model performance.

\begin{table}[h]
    \centering
    \begin{tabular}{ccccc}
        \toprule
        \textbf{Model} & \textbf{Feature} &\textbf{Error} & \textbf{Percentage} & \textbf{F1 Score} \\
        \midrule
        lda & all & duplicated & 50 & 0.9675 \\
        lda & all &duplicated & 30 & 0.9626 \\
        \dots&  \dots &  \dots &  \dots &  \dots \\
        et  & all &duplicated & 30 & 0.9396 \\
        lr  & person\_age    & outlier &30 & 0.9378 \\
        lr  & cb\_person\_cred... & outlier &50 & 0.9374 \\
        lr  & cb\_person\_cred... & outlier & 30& 0.9369 \\
        lda & cb\_person\_def...  & noise & 30& 0.9367 \\
        lda & person\_emp\_...  & noise & 50& 0.9366 \\
        \bottomrule
    \end{tabular}
    \caption{Top performance of ML Model, for each Feature, Error and percentage.}
    \label{tab:top_model}
\end{table}

\section{Conclusion and Future Work}
\label{sec:conclusion}
Assessing credit risk is a vital function essential for the effective and prudent management of financial institutions. With advancements in artificial intelligence, credit risk models can now leverage machine learning algorithms. These models necessitate large datasets for training, with their accuracy hinging critically on the input data quality.

In this study, we explored the impact of various data quality issues within the training dataset on the effectiveness of several machine learning algorithms. Interestingly, our findings indicate that certain errors—like outliers, missing values, and duplicates—can sometimes enhance model performance in specific cases. Conversely, imperfections such as noise and mislabeling generally lead to a significant decline in model performance.

Our investigation demonstrates that the consequences of data quality issues are influenced by the particular feature affected, the architecture of the model, and the type of error introduced. To bolster this analysis, we present a methodology for systematically perturbing datasets using the PuckTrick library, which allows for the controlled introduction of diverse error types and supports a thorough evaluation of model sensitivity to data flaws.

Possible future works include applying the same methodology to different credit datasets. Additionally, extending our analysis to cover not only binary problems but also multiclass tasks and regression scenarios is another direction. Finally, investigating the impact of data corruption on fairness presents an intriguing avenue for further research.

\bibliographystyle{ACM-Reference-Format}
\bibliography{biblio}

\end{document}